\pdfoutput=1
\documentclass[10pt, a4paper]{article}

\usepackage[]{lrec-coling2024} 

\usepackage{times}
\usepackage{array}
\usepackage{setspace}
\usepackage{latexsym}
\usepackage{booktabs}
\usepackage{caption}
\usepackage{bm}
\usepackage{multirow}
\usepackage{enumitem}
\usepackage{float}
\usepackage{graphicx}
\usepackage{subcaption}
\usepackage{xcolor}
\usepackage{soul}
\usepackage{titlesec}
\usepackage{balance}
\usepackage[T1]{fontenc}
\usepackage[utf8]{inputenc}
\usepackage{microtype}
\usepackage{inconsolata}
\usepackage{amsmath}
\usepackage{adjustbox}
\usepackage{tipa}
\usepackage{multicol,dblfloatfix}

\title{Aspect-oriented Consumer Health Answer Summarization}

\name{Rochana Chaturvedi, Abari Bhattacharya, Shweta Yadav} 

\address{University of Illinois Chicago \\
         Chicago, Illinois, USA\\
         \{rchatu2, abhatt62, shwetay\}@uic.org\\}

\abstract{Community Question-Answering (CQA) forums have revolutionized how people seek information, especially those related to their healthcare needs, placing their trust in the collective wisdom of the public. However, there can be several answers in response to a single query, which makes it hard to grasp the key information related to the specific health concern. Typically, CQA forums feature a single top-voted answer as a representative summary for each query. However, a single answer overlooks the alternative solutions and other information frequently offered in other responses. Our research focuses on aspect-based summarization of health answers to address this limitation. Summarization of responses under different aspects such as suggestions, information, personal experiences, and questions can enhance the usability of the platforms. We formalize a multi-stage annotation guideline and contribute a unique dataset comprising aspect-based human-written health answer summaries. We build an automated multi-faceted answer summarization pipeline with this dataset based on task-specific fine-tuning of several state-of-the-art models. The pipeline leverages question similarity to retrieve relevant answer sentences, subsequently classifying them into the appropriate aspect type. Following this, we employ several recent abstractive summarization models to generate aspect-based summaries. Finally, we present a comprehensive human analysis and find that our summaries rank high in capturing relevant content and a wide range of solutions. 
 \\ \newline 
 }
 
\begin{document}

\maketitleabstract
\section{Introduction}
Over the last decade, Community Question Answering (CQA) forums such as Yahoo! Answers, Stack Exchange, and Reddit have become popular platforms for individuals to seek information. An estimated 79\% of Internet users search for health information online, and 74\% of such users turn to social media-based platforms facilitating peer-to-peer healthcare \citep{wang2023health}. Even though these platforms are not necessarily frequented by experts, people often turn to them for their health-related inquiries for reasons such as easily and freely available information, avoiding medical jargon, inhibition in discussing sensitive personal information in person, distrust in modern medicine, and learning from first-hand experiences of others \citep{firstHandExperience}. In addition, access to medical experts may be limited and may be prohibitively expensive \citep{baeten2018inequalities}. However, being an informal setup with minimal regulations, the pertinent information becomes entangled in lengthy, convoluted responses plagued by redundancy and irrelevant content such as sarcasm, humor, or off-topic discussions. These observations indicate a need for answer summarization in the community health question-answering domain, which directly impacts the well-being of people.%\citep{kaminski2020analysis}

Several works on CQA answer summarization focus on a single best-voted answer \citep{chowdhury2019cqasumm, chowdhury2021neural}, or multi-QA pair summarization \citep{hsu2022summarizing}, pairing a single answer with a query. However, in line with multi-answer summarization works \citep{liu2008understanding, fabbri2022answersumm}, we find that a single answer need not be the unique correct answer. Additionally, structured presentation of the information into aspect-driven summaries can be more useful for the end users \citep{tauchmann2018beyond, frermann2019inducing}. We identify four essential answer aspects in our data\textemdash (Clarificatory) Questions, Information, Suggestions, and (Personal) Experiences. We are motivated by the observations that the users are interested in separating facts (information) from opinions (suggestion, experiences) \citep{yu2003towards}, suggestions from information \citep{negi2015towards}, extracting personal experiences \citep{hedegaard2013extracting}, and identifying unresolved discussions \citep{kim2014towards}, one way of which is identifying clarificatory questions. 

\begin{figure}[t]
    \centering
\includegraphics[width=0.48\textwidth, height=6.4cm]{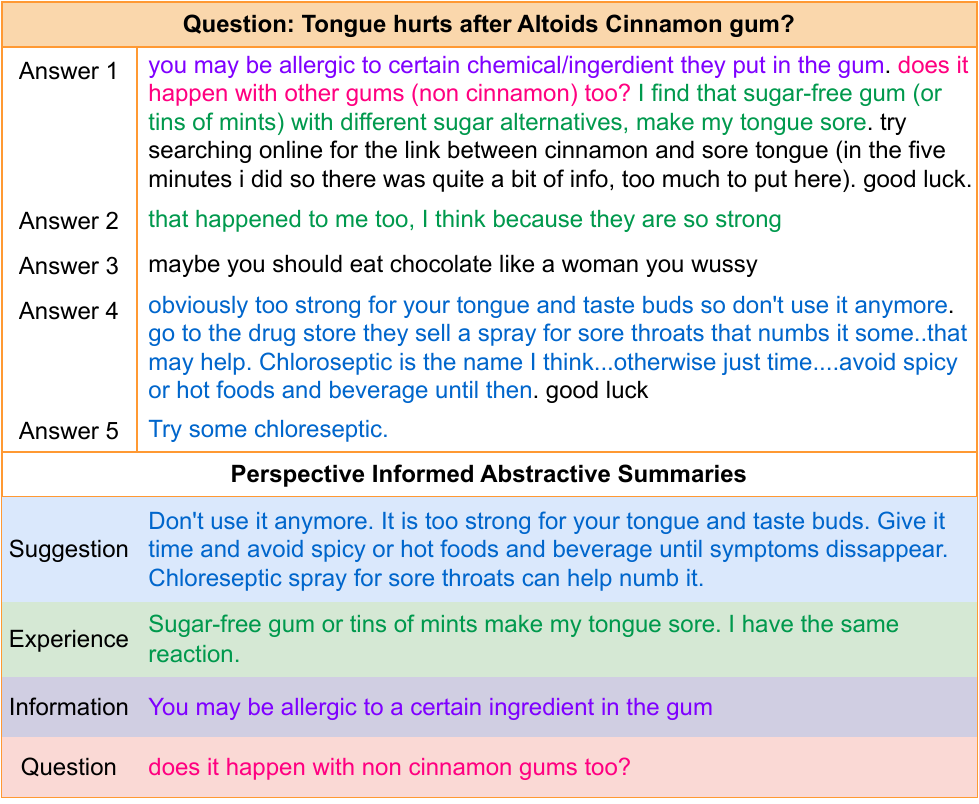}
\setlength{\belowcaptionskip}{-14pt}
\caption{Example illustrating a health query, responses, and their aspect-based summaries.}
    \label{fig:example}
\end{figure}

Consider the example in Figure \ref{fig:example}, where part of Answer 1 contains \textit{Information} on the cause of the problem, part of it shares a \textit{Personal Experience} concerning the query, and part of it raises a \textit{Clarificatory Question} to obtain information about the precise context of the problem. Answer 2 contains another first-hand experience, and answers 4 and 5 provide some \textit{Suggestion} to alleviate the problem. Answer 3 is an insult, while part of Answer 1 (in black) does not add any information. Therefore, both of these are irrelevant.

We contribute:
\textbf{(1)} CHA-Summ: Consumer Health Answers Multi-Aspect Summarization dataset\textemdash the first summarization dataset catering to health answers in the CQA forums. The data is constructed using a multi-step annotation framework. We perform sentence-level annotations for relevance classification and aspect classification and finally provide human-written answer summaries across the four aspects. %add data size range 4.5K sentences across 210 threads.
\textbf{(2)} We build an automated multi-step summarization framework based on several pre-trained transformer-based models which have enabled the generation of concise, human-like summaries \citep{zhang2301benchmarking, yang2023exploring}. Our pipeline selects answer sentences relevant to the query, classifies them into an aspect category, and generates aspect-based abstractive summaries.
\textbf{(3)} We perform a comprehensive human evaluation of system-generated summaries and find that our summaries cover a wide range of relevant information under appropriate aspect types and identify key areas for future research.

\section{Related Works}
 Multi-answer summarization can be modeled as multi-document query-focused summarization \citep{dang2005overview,daume2006bayesian}. This approach eliminates extraneous answer sentences (documents) not aligned with the query. Typically, such systems produce a single summary, assuming a lack of structure. We venture beyond this and create categorical summaries across different pre-defined aspect types in the source answers. Previous works in aspect-based summarization define fine-grained aspects in specific domains, such as product attributes from the reviews in the e-commerce domain \citep{gerani2019modeling, yang2018aspect}, or discussion topics in news articles \citep{frermann2019inducing}, or define aspects over multiple domains \citep{hayashi2021wikiasp}. Our work focuses on pre-determined aspects in health-related answers on CQA forums. %Usually, these aspects are not globally defined and are determined separately for each thread \citep{fabbri2022answersumm}. In contrast, our work focuses on broad, pre-determined aspects shared across all question threads.

The existing works catering to summarization in the biomedical domain focus on single-document summarization of clinical records \citep{zhang2020optimizing, macavaney2019ontology, hu2022graph}, biomedical literature \citep{zhang2020pegasus}, multi-document summarization of clinical trials \citep{wallace2021generating}, biomedical literature \citep{deyoung2021ms2, cohan2018discourse}, expert-sourced medical answers \citep{abacha2019summarization, zhang2020summarizing}, etc. In contrast to expert-sourced answers, the information on CQA forums is noisy and characterized by typos, grammatical oversights, and informal language. There is limited work at the intersection of noisy CQA forums and the biomedical domain. Here \citet{savery2020question} introduce the task of medical question-driven answer summarization, and several works focus on question summarization \citep{yadav2022towards,yadav2022question,yadav2021reinforcement}. We add to this literature by venturing beyond question summarization to the task of health answer summarization in the CQA domain, which can help address inequity in access to health-related information. 

\section{Data Creation}
We begin with the Yahoo!L6 dataset containing anonymized questions from Yahoo!Answers across a variety of topics and their answers.\footnote{ The data can be downloaded from \url{https://webscope.sandbox.yahoo.com/}.} The dataset comprises several other fields such as category, number of answers, best answer, date, and language. We select only English language question-answer threads. Since our focus is on the healthcare domain, we further retain the threads specific to the Health category, including $21$ sub-categories such as Allergies, Diabetes, Heart Diseases, etc.\footnote{See Appendix \ref{appendix:data}, Figure \ref{fig:subcat-distribution} Panel (a) for sub-categories distribution.} 

\paragraph{Sampling Strategy} 
We first sample threads based on the number of answers, which can range from zero to 2235 in response to a single query.\footnote{Distributions of the number of answers for each sub-category are presented in Appendix \ref{appendix:data} Figure \ref{fig:subcat-distribution} Panel (b).} We use Tukey Fence criteria \citep{tukey1977exploratory} to eliminate outliers. This popular non-parametric outlier detection heuristic creates upper and lower `fences' (or boundaries) at a distance of 1.5 interquartile range (distance between first and third quartiles) before the first and beyond the third quartile. Any data points outside these fences are considered outliers. Thereafter, the data contains 4\textendash6 answers on average across all categories. We retain all threads with the number of answers in this range, resulting in a subset of around 77K question-answer threads. 

From this, we randomly subsample $10$ threads to annotate for each sub-category, resulting in $210$ threads containing more than $4000$ sentences. The data is noisy and comprises some irrelevant queries, especially in the more generic sub-categories like Other-Health and Other-Health \& Beauty. Therefore, we review each thread to ensure that our data consists of queries on disease or condition, drug or treatment, medical diagnosis, tests and therapeutic procedures, anatomy, medical organizations, experiences, or any other topic related to personal or public health. We add human-written summaries for answers provided in each thread to produce our gold-standard dataset as described next.
% provide their description in Appendix \ref{appendix:perspectives}.\footnote{Our initial aspect types are inspired from question types defined in \citet{yadav2023towards}.}  we identify all available answer categories in our data and To make the problem tractable, we merge some of the rare answer categories in the data under broader classes
\paragraph{Annotation Process}
To identify the aspect categories, we begin with question types defined in \citep{yadav2023towards}. After analyzing the answer sentences in our data, we find the following categories to be present\textemdash \textit{Information, Causes, Testing, Treatment, Symptom, Diagnosis, Drug, Complication, Side-effects}, and \textit{Duration}. Following \citep{bhattacharya2022lchqa}, we further add the categories \textit{Suggestion}, as people may provide alternatives to modern medicine, \textit{Experience} for personal accounts shared by users, and \textit{Question} that seek further clarification. We also find the \textit{Emotional Support} category for sentences such as ``you got this!'',  ``don’t worry''. However, we omit this aspect as such sentences are highly repetitive and irrelevant to information summarization. Moreover, very few people are soliciting emotional support from these forums \citep{kaminski2020analysis}. We also find some of the other answer categories to be quite rare. These are \textit{Causes, Testing, Symptom, Diagnosis, Complication, Side-effects}, and \textit{Duration}. We merge these with the broader \textit{Information} category. We also merge the rare categories \textit{Drug} and \textit{Treatment} suggestions under the broad aspect \textit{Suggestion}. For example, ``Have ginger tea and Benadryl to soothe cough.''. Our final four aspect types are \textemdash \textit{Suggestion}, \textit{Experience}, \textit{Information}, and \textit{Question}. We define these in Table \ref{tab:finalperspectivedefinition}. We pre-process the answer sentences and perform sentence tokenization.\footnote{See Appendix \ref{appendix:preprocessing} for details of pre-processing and tokenization.} Then, two annotators with excellent English proficiency and a medical informatics background write aspect-based answer summaries in the following phases.\footnote{Each annotator completes all phases for a thread on their own for consistency.}

\begin{table*}[htbp]
    \centering
    \caption{Final Answer Aspects\textemdash Definitions and Examples. Examples comprise a question and an answer sentence with the given aspect.}
    \resizebox{1\linewidth}{!}{
    \begin{tabular}{p{1.7cm}p{15.3cm}p{10cm}}
    \toprule
   \textbf{Aspect}& \textbf{Description}&\textbf{Example:Question, Answer Aspect} \\
   \midrule

\textit{Suggestion}& A recommendation or advice such as medical or therapeutic interventions, including home remedies, referral to a care provider, etc., aimed at addressing a health concern or preventing a health issue from arising, exacerbating, or alleviating the symptoms. & \vspace{-\abovedisplayskip}\smallskip\begin{minipage}{\linewidth} Q. How can I relieve TENDONITIS in my shoulder??	\\A. Your doctor can give you a cortisone injection directly to the site of the pain to decrease the inflammation. \end{minipage}\\\midrule

\textit{Experience}& People often share their personal experiences in support of the query. These may contain valuable subjective insights about what did or did not work for those in the same situation. Perhaps this is the most important aspect that attracts people to public forums for their health-related concerns. & \vspace{-\abovedisplayskip}\smallskip\begin{minipage}{\linewidth}  Q. Newly diagnosed with sleep apnea - please advice!?	\\A. mine is treated with a machine which i wear while i sleep.\end{minipage} \\\midrule
  
 \textit{Information}& Any knowledge regarding various aspects of healthcare, such as epidemiology of a disease, its root cause, whether it is progressive, how effective a certain medicine is, if it has any side effects, what is entailed in a particular medical procedure, any adverse conditions that may arise from it or as a result of pre-existing medical conditions, general statistics about public health issues in a particular country etc. & \vspace{-\abovedisplayskip}\smallskip\begin{minipage}{\linewidth} Q. what is coumadin?	\\A. It is used for the prophylaxis of thrombosis and embolism in many disorders.\end{minipage}
 \\ \midrule
\textit{Question} & In several cases, the original question may miss out on important details to enable accurate answering. In such cases, the answer providers may seek further clarification. &\vspace{-\abovedisplayskip}\smallskip\begin{minipage}{\linewidth}  Q. How can I find one of the best neurologists in the United States?	\\A. What is your condition?\end{minipage}\\
\bottomrule
 \end{tabular}}

 \label{tab:finalperspectivedefinition}
\end{table*}

\begin{enumerate} [noitemsep]
\item Identify relevant answer sentences: Public platforms might encourage a very informal setup, and people often post humorous replies or digress to some other subject in their answers. For instance, given the question `What could potentially happen after having a stroke?', the answer sentence 'You can even look at some older people and say here is someone who is 20\% dead' does not add any meaningful contribution. Further, as discussed before, the sentences that provide emotional support, although relevant to the question, are irrelevant for summarization. Therefore, we classify such answer sentences as irrelevant to the question and answers directly related to the question are classified as relevant. 
\item In this round, we classify all the relevant sentences into one of our final four aspects. For example, in Figure \ref{fig:example}, one answer sentence \textit{suggests} using Chloreseptic spray as a remedy while another shares a personal \textit{Experience} having the same reaction with sugar-free gums.
\item In this round, we write abstractive answer summaries for each of the four categories as applicable. Note that a particular aspect might not appear in every thread.
\end{enumerate}
To compute inter-annotator agreement (IAA) for the first phase, we randomly sample 200 question-sentence pairs and annotate them for relevant sentence identification. We obtain a Cohen's Kappa of 0.64. For the classification task, we sample and annotate 200 relevant sentences into appropriate aspects and achieve a Cohen's Kappa of 0.63. Both the scores indicate substantial agreement. 

\subsection{Analysis}
We report the final number of observations for each aspect type in Table \ref{tab:statistics}. The number of aspect categories in an observation ranges from 1 to 4, with 2 being the average. We also find that five threads have no irrelevant sentences among the provided answers.

\begin{table}[]
\centering
\resizebox{0.8\linewidth}{!}{%
\begin{tabular}{lccc}
\toprule
\textbf{Relevance} & \textbf{Aspect} & \textbf{\#threads} & \textbf{\#Sentences} \\ \midrule
\multirow{4}{*}{Relevant}&Suggestion & 162 & 1202\\
&Experience & 112 & 426\\
&Information& 163 & 1199\\
&\textit{Question} & 29& 59\\
Irrelevant & none & 205 & 1532\\
\bottomrule
\end{tabular}
}  
% \end{adjustbox}
\caption{Number of threads and the number of sentences associated with each aspect across the CHA-Summ dataset. The dataset consists of 4418 sentences across 210 answer threads.}  
\label{tab:statistics}
\vspace{-1em}
\end{table}

\begin{figure}[ht]
    \centering
    \includegraphics[width=0.45\textwidth]{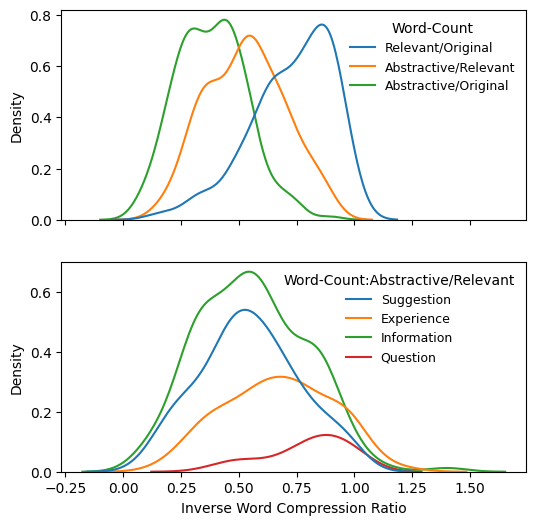}
\caption{Distribution of word-count ratio. The top panel shows the compression over each annotation step. The Bottom Panel shows the reduction for each aspect category in the final summaries.}
    \label{fig:compression}
\end{figure}

  We show the reduced word count distribution or the inverse compression ratio in Figure \ref{fig:compression}. When comparing the number of words in the original answer to those in the final summaries combined across aspect types, we observe a reduction to 38\% on average, with a standard deviation of 0.15\%.\footnote{Appendix \ref{appendix:compression_health_cat} Figure \ref{fig:compression_cat} reports reduction across health categories. The categories with the highest reduction are  Respiratory Diseases, Men's Health, and Diet and Fitness. While those with the least reduction are Other-General Healthcare, Other-Health \& Beauty, and Alternative Medicine.} In our annotation pipeline, we observe a reduction to 73\% of the original length on average (with a standard deviation of 18\%) just by weeding out irrelevant sentences. In five threads, even though all answer sentences are identified as relevant, we still achieve some compression while summarizing them. We also analyze the reduction across each aspect type by taking the ratio of the number of words in the final aspect-specific summary to the number of words in the relevant sentences classified under that aspect. The biggest average reduction is for the \textit{Suggestion} to 55\%, and the smallest average reduction is for \textit{Question} to 80\%, and this is also the most sparse class. In a few examples, a particular aspect summary is longer than the relevant sentence cluster in that category. E.g. ``6 wks to heal.'' is summarized to a complete sentence ``It takes 6 weeks to heal.''. ``I'm trying YOU on a Diet  Dr. Oz.'' to ``I'm trying "YOU on a Diet" by Dr. Oz''. ``There is no medicine to "kill" hair.'' to ``There is no medicine to remove hair permanently.'' 

 Finally, we shuffle our data and split it into train-val-test at the thread level (or final summaries level) in the ratio 60:20:20 stratified by health categories.
 
\section{Summarization Pipeline}
To benchmark our dataset,
we devise the following pipeline to generate aspect-based summaries of health answers from a community QA forum.
\paragraph{Task Definition}
Given a natural language health query $q$ and a number ($n$) of answer sentences $A = \{a_1, a_2, ...a_n\}, n\ge1$ provided by different users, we learn the following mapping:\\
(1) $r:(A, q)\mapsto\{0,1\}$ where $r_{a,q}=1$ if the sentence $a\in A$ is relevant to $q$. \\
(2) $f:A_r\mapsto C, A_r\subseteq A, r_{a,q}=1 \forall a\in A_r$, mapping all relevant sentences to the most appropriate aspect type $c\in C=\{Suggestion, Experience, Information, Question\}$. \\
(3) Generate aspect-based summaries $\tilde{A_c}, \forall c\in C$, $Summ_{c}: A_c\mapsto \tilde{A_c}, A_c\subseteq \{a|(f_a=c)\wedge (r_{a,q}=1), a\in A\}$ 

\subsection {Relevant sentence Selection: $r:(A, q)\mapsto {0,1}$}
For this subtask, our goal is to find all answer sentences that are relevant to the query. We experiment with several transformer-based language models for sentence-pair classification:  
\paragraph{SBERT} We hypothesize query and relevant answers to be semantically more similar than query and irrelevant sentences. We use Sentence-BERT (SBERT) \citep{sentence-bert} embeddings and compute the cosine similarity between a question and an answer sentence. The model is pre-trained with sentence-similarity relation as the objective.  We experiment with several pre-trained variants and find that `all-mpnet-base-v2' embeddings provide the best separation between the relevant and irrelevant labels on the training set. Finally, we train a logistic regression (LR) model to classify cosine similarity scores into one of the relevant/irrelevant labels. We use balanced class weights and 10-fold CV with grid search to choose the regularization parameter and report the evaluation results on the test split.
\paragraph{SBERT-ft}
Fine-tune a Sentence BERT (SBERT) using triplet loss. We construct 17919 triplets $\{(q, a, a')|r_{a,q}=1\wedge r_{a',q}=0\}$ from the training set by using a question ($q$) as an anchor and all possible combinations of relevant ($a$) and irrelevant ($a'$) answer sentences to that question as positive and negative samples respectively.\footnote{We exclude four training and one validation set threads with no irrelevant sentence for triplet creation.} We warm start with all-mpnet-base-v2 for 1 epoch and fine-tune the model for 10 epochs. %\footnote{We also experimented with distilroberta which has worse performance.} 
We save the best model using triplet evaluation on the validation set. Finally, we compute the cosine similarity between the embeddings, train an LR model as above, and report the results on test set. 

\paragraph{BERT for sentence-pair classification} In this experiment, we fine-tune the BERT variants for sequence classification by combining question and answer pairs with a separator token. We fine-tune the model for 5 epochs and save the model with the best validation loss. We experiment with BERT-base-cased, BioMedRoBERTa-base, and RoBERTa-base. We use a learning rate of 2e-5 with weight decay by 0.01 fraction after 100 warmup steps. %adding a balanced loss function doesn't help here. 

\subsection{Aspect Classification: $f:A_r\mapsto C$} In this step, we classify the retrieved sentences under different aspects. 
\paragraph{Zero-shot Classification (ZS)} We begin with a zero-shot pipeline (ZS) using a prompt and predict paradigm \citep{yin2019benchmarking} based on NLI. This is a popular approach for zero-shot inference from a large pre-trained model. Here, we prepare a prompting template, where each candidate label is posed as a hypothesis and the sequence to classify is posed as the premise in the format: \textit{<cls>Premise<sep>Hypothesis<sep>}. For example: \textit{<cls>If you have fallen arches you may want to get arch supports for your shoes<sep>This example is treatment<sep>}.
\noindent Scores for entailment are passed through a softmax classifier to obtain target classification. We choose the joeddav/bart-large-mnli-yahoo-answers model trained on multiNLI (MNLI) dataset \citep{williams-etal-2018-broad} and fine-tuned on the Yahoo!Answers topic classification. We use the following candidate labels\textemdash `informative', `information', `cause', `question', `interrogative', `suggestion', `imperative', `instruction', `command', `personal experience', `experience', `personal'. If the output label is `informative', `information', or `cause', we classify the sentence under \textit{Information}; if it is `question' or `interrogative' we classify it as \textit{Question}; if `suggestion', `imperative', `instruction', `command', as \textit{Suggestion}; otherwise as \textit{Experience}. 
\paragraph{Linguistic Knowledge-enhanced Classifiers}
We acquire certain task-specific linguistic heuristics that can potentially improve over the simple zero-shot baseline. 

(1) \textit{Personal Pronouns (PP)}: We observe that the \textit{Experience} category often consists of sentences with personal pronouns. E.g. `My mother had orthodontia done in her 50s with no problems.'

(2) \textit{Grammatical Moods (GM)}: the other aspect classes have a close correspondence with the grammatical moods in the English language. The sentences corresponding to \textit{Suggestion} have an `imperative' mood. E.g. `Try essential oils'. Those from \textit{Information} category have an `indicative' mood. E.g. `Thyroid cancer is very slow growing'. Those of type \textit{Question} have an `interrogative' mood. E.g. `Are you low on vitamin B12?'.

We propose the following variants to leverage these heuristics:
\begin{enumerate}
    \item \textbf{ZS+PP}: Here we leverage only the first heuristic. We follow the same procedure as the baseline zero-shot approach except when the output of zero-shot classification is `informative', `information', or `cause', we first check if the sentence contains personal pronouns (PP) and if so, classify it as \textit{Experience} otherwise we classify it as \textit{Information}.

    \item \textbf{GM Classifier}
    We train a feature-based grammatical moods (GM) model to classify each sentence into an aspect. The GM model is a Logistic Regression classifier trained using 10-fold CV and balanced class weights with the following features:\\
    \textit{(a) Moods Probabilities}: To obtain these, we train an initial moods classifier by augmenting input with  Parts-Of-Speech (POS) tags, following the best model from \citet{lettergram}. The classifier is a Convolutional Neural Network (CNN) \citep{lecun1989backpropagation} model trained with a dataset comprising interrogative sentences from SQuAD 2.0 \citep{rajpurkar2018know} and interrogatives, imperatives and indicative sentences from Speech Act Annotated Dialogues (SPAADIA) corpus \citep{leech2013spaadia} along with additional human crafted imperative sentences. This data is then split into training and test sets in the ratio 80:20. The model achieves an impressive 98.94\% macro-F1 score over the 20\% test split. The details of the training process are provided in Appendix \ref{appendix:moodsclf}. 
    For inference on our dataset, we process each sentence in our data using this model and obtain imperative, interrogative, and indicative probabilities. \\
    \textit{(b) Personal Pronoun (PP)}: a binary feature marked true if the sentence contains a personal pronoun and false otherwise.\\
    \textit{(c) Question Features}: We use several additional binary features related to question identification\textemdash question\_mark (true if the sentence ends in a `?'), do\_pattern (true if the sentence contains patterns such as ``do i'', ``are there'', ``tell me more'', etc.),  helping\_verb (true if the sentence starts with helping verbs such as `is', `are', `will', etc.).\footnote{The complete list for do\_patterns and helping\_verbs is provided in Appendix \ref{Appendix:ques_features}.}   
    
\end{enumerate}
   
\paragraph{Fine-tuned RoBERTa}
Finally, we also fine-tune RoBERTa for multi-class classification of an answer sentence into one of the four aspects. We use the roberta-base and fine-tune the model for 10 epochs while saving the model checkpoint with minimum validation loss. We start with a learning rate of 5e-5 for 50 warm-up steps and then a gradual weight decay by a factor of 0.01.

\subsection{Aspect-based Answer Summarization: $Summ_{c}: A_c\mapsto \tilde{A_c}$}
\label{Summary - Method}
In this step, we first chunk together all the relevant sentences across a particular aspect class in the order they were in the source answers. Then, we summarize them using transformer-based state-of-the-art abstractive summarization models. In particular, we experiment with BART \citep{lewis-etal-2020-bart}, which uses a denoising auto-encoder trained to regenerate arbitrarily corrupted text.
T5 \citep{t5}, a model that generalizes text-to-text transfer learning to a variety of NLP tasks, trained by randomly corrupting text spans.  Prophetnet \citep{qi2020prophetnet} trained with a novel self-supervised objective of future n-gram prediction, along with a new n-stream self-attention mechanism.
And, Pegasus \citep{zhang2020pegasus}, an encoder-decoder architecture trained with a novel self-supervised objective of Gap Sentence Generation by masking full sentences instead of random tokens or text spans, designed especially for abstractive summarization.
% \footnote{We use the pre-trained variants ainize/bart-base-cnn, t5-small,, microsoft/prophetnet-large-uncased, and google/pegasus-large from the transformers library \citep{wolf-etal-2020-transformers}}
We save a separate model for each aspect type. We fine-tune the models under the following two scenarios\footnote{See Appendix \ref{appendix:summary generation} for implementation details.}:\\
\textbf{Ans+ft}: We use the source answer as the input to fine-tune the model and to obtain the final summaries. We want to test if the models can identify the answer categories by looking at the target summary for the respective aspect.\\
\textbf{Pipeline+ft}: We use the gold annotations of the relevant sentences under a particular aspect class (over-extractive) for fine-tuning the model. These are the outputs of our second round of annotations. At the time of inference, we use system-generated relevant sentences in a particular category from the test set as the input. These are the output from the second step of our pipeline $A_p$.
 
\section{Result Analysis}
\subsection{Experimental Results}
\paragraph{Relevant Sentence Selection $(r)$:} The results from experiments on this subtask are reported in Table \ref{tab:relevant}. While the cosine-similarity classifier using the off-the-shelf Sentence BERT model does quite well, fine-tuning it provides an improvement of more than 1\%, and fine-tuning a RoBERTa model provides more than 2\% improvement in the macro-F1 score. 
\begin{table}[htbp]
\centering
\caption{Evaluation results for Relevant Sentence Selection on the held-out test set.} 
% \begin{adjustbox}{max width=.44\textwidth}
% \scriptsize
\resizebox{0.9\linewidth}{!}{%
\begin{tabular}{lccc}
\toprule
\multirow{2}{*}{\textbf{Approach}} & \multicolumn{3}{c}{\textbf{$F_1$}}\\\cline{2-4}
 &\textbf{Relevant} & \textbf{Irrelevant} &\textbf{Macro-avg}\\ \midrule
SBERT$_r$ (baseline)&74.53 &63.75 & 69.14\\
SBERT-ft$_r$ & 79.39 &61.31 & 70.35\\
BERT-ft& 60.65 & 77.78& 69.21\\ 
BioMedRoBERTa-ft & 64.16 &  78.78& 71.47 \\
RoBERTa-ft$_r$ &78.74 &64.27 & 71.51\\%3.10\\%71 &

\midrule
\#Sentences & 525 & 334& 859\\
\bottomrule
\end{tabular}
}
% \end{adjustbox}
\setlength{\belowcaptionskip}{-2pt} 
\label{tab:relevant}
\vspace{-1em}
\end{table}

\begin{figure}[ht]
  \centering
\includegraphics[width=0.4\textwidth]{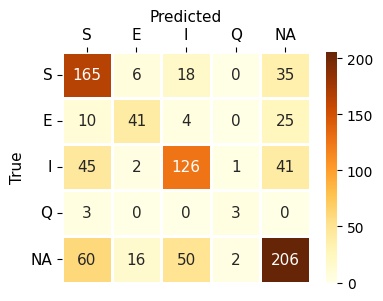}

  \setlength{\belowcaptionskip}{-12pt}\caption{Confusion Matrix for end-to-end aspect identification using RoBERTa-ft$_r$ for selecting relevant sentences, RoBERTa-ft$_f$ for aspect classification. \textbf{S}uggestion, \textbf{E}xperience, \textbf{I}nformation, \textbf{Q}uestion), NA (irrelevant).
  }
  \label{fig:perspective_identification_cm}
\end{figure}

\paragraph{Aspect Identification $(f)$:} 
We report the benchmarking results for this task in Table \ref{tab:perspective_classification}. We find that an NLI-based zero-shot (ZS) classifier's performance over \textit{Suggestion} and \textit{Experience} classes is quite poor. Combining this classifier with personal pronoun heuristic in ZS+PP improves the $F_1$ score for \textit{Experience} category by 20\% and provides an improvement in the macro-$F_1$ by around 5.5\%. Leveraging both personal pronoun and grammatical mood heuristics in the GM classifier gives an overall improvement of almost 24\% over the ZS baseline. Finally, we observe that fine-tuning RoBERTa surpasses even this and gives us a remarkable 37\% improvement over the ZS baseline.
For end-to-end aspect identification, we first take the relevant sentences classified by RoBERTa-ft$_r$ for relevant sentence classification and assign the aspect type assigned by RoBERTa-ft$_f$ to these. While we assign 'none' to all sentences classified as irrelevant. The overall combined macro-average $F_1$ score is 59.39\%. We provide the confusion matrix for this end-to-end task in Figure \ref{fig:perspective_identification_cm}. We note that most errors are cascaded from the relevant sentence selection module.

\begin{table*}[htbp]
\centering
\begin{adjustbox}{max width=0.9\textwidth}
\begin{tabular}{l ccc ccc ccc ccc ccc ccc c}
\toprule
 \multirow{2}{*}{\textbf{Approach}} & \multicolumn{3}{c }{\textbf{Suggestion} }& \multicolumn{3}{c}{\textbf{Experience}} & \multicolumn{3}{c}{\textbf{Information}} & \multicolumn{3}{c}{\textbf{Question}} & \textbf{Avg}\\
 \cmidrule(r){2-14}
 
 & \textbf{P} & \textbf{R} & \bm{$F_1$} %sugg
 & \textbf{P} & \textbf{R}& \bm{$F_1$} %exp
 & \textbf{P} & \textbf{R} & \bm{$F_1$} %info
 & \textbf{P} & \textbf{R}& \bm{$F_1$} %clari
 & \bm{$F_1$}\\
 \midrule
ZS   &88.24 & 6.70 &  12.45 %sugg
  &23.62 &  37.50 & 28.99  %exp
  &41.38  & 72.56 & 52.70 %info
  &50.00  & 33.33 &  40.00  %clar
  & 33.53\\
ZS+PP
& 88.24  & 06.70 &  12.45
& 34.20 &  82.50 &  48.35
& 47.27 &  68.37 &  55.89
&50.00 &  33.33 &  40.00 
&39.17\\

GM
& 72.73 &  42.86  & 53.93%sugg
& 62.77 & 73.75 & 67.82 %exp
& 59.31 & 80.00 & 68.12 %info
& 33.33 & 50.00 & 40.00 %que
& 57.47  \\
RoBERTa-ft$_f$
& 71.21 & 83.93 & 77.05%sugg
& 79.45 &  72.50 &  75.82 %exp
& 81.97 &  69.77 &  75.38 %info
& 60.00 &  50.00 &  54.55 %que
& 70.70 \\

\midrule
Observations & \multicolumn{3}{c}{224} &\multicolumn{3}{c}{80} &\multicolumn{3}{c}{215} &\multicolumn{3}{c}{6} & 525 \\
 
\bottomrule
\end{tabular}
\end{adjustbox}

\caption{Evaluation results for the subtask of aspect classification of sentences on the held-out test set. The table reports macro-average $F_1$ scores.}
\label{tab:perspective_classification}
\end{table*}
 
\paragraph{Aspect-Based Answer Summarization $(Summ_c)$:} 
Considering that only the aspects that appear in original answers would have a corresponding gold summary, we pair together the system and gold summaries for an aspect and evaluate them using ROUGE scores \citep{lin2004rouge} reported in Table \ref{tabSummaries}. We observe that across all model variants, our pipeline, in general, provides significant improvements in comparison to the respective Ans+ft variants. The improvement can be as high as 31\% in terms of ROUGE-L, for example by fine-tuning Pegasus over \textit{Experience} category. Averaging across models and the aspect types (excluding \textit{Question}), there's an improvement of around 10\% in terms of ROUGE-1 and ROUGE-L, and 7\% in terms of ROUGE-2 from the pipeline strategy over Ans+ft strategy. We also find that fine-tuning using the source answers is not an effective strategy for the sparse \textit{Question} category where none of the models in the second panel produce any summaries while looking at the bottom panel of the Table, BART and Pegasus perform exceptionally well using the pipeline strategy. We also consider the standard best answer summary for comparison and find that although it is similar in performance to Ans+ft variants over the \textit{Suggestion} and \textit{Information} categories and better on the \textit{Question} category, it is quite poor in the \textit{Experience} category and worse than Pipeline+ft for all categories. Overall, the BART model fine-tuned using the pipeline strategy has strong performance across all categories.
\begin{table*}[ht]
\centering
\caption{ROUGE 1/2/L scores for held-out test set for end-to-end aspect-based summarization. We evaluate each aspect using threads where the original answer contains that aspect.}
\begin{adjustbox}{max width=0.8\textwidth}
\begin{tabular}{cl ccc ccc ccc ccc ccc ccc} \toprule
 % Approach
\multirow{2}{*}{} &\multirow{2}{*}{\textbf{Model}} &\multicolumn{3}{c}{\textbf{Suggestion}} & \multicolumn{3}{c}{\textbf{Experience}} & \multicolumn{3}{c}{\textbf{Information}} & \multicolumn{3}{c}{\textbf{Question}}\\
 \cmidrule(r){3-14}
 && \textbf{R1}&\textbf{R2}&\textbf{RL}& \textbf{R1}&\textbf{R2}&\textbf{RL}& \textbf{R1}&\textbf{R2}&\textbf{RL}& \textbf{R1}&\textbf{R2}&\textbf{RL}\\
 \midrule
& Best Ans &0.32 & 0.29 & 0.27&
             0.16 &0.08 &0.12&
              0.35&0.32&0.31&
              0.16&0.02&0.10\\
\midrule
\parbox[t]{2mm}{\multirow{4}{*}{\rotatebox[origin=c]{90}{\small Ans + ft}}} & BART&0.33 & 0.27 & 0.29&
              0.39&0.41&0.37&
              0.36&0.32&0.32&
              0.0&0.0&0.0\\
 &T5& 0.34 & 0.20 & 0.28&
              0.37 & 0.38 & 0.34 &
              0.37 & 0.31 & 0.32 &
              0.0 & 0.0 & 0.0\\
 &Pegasus& 0.34 & 0.26 & 0.26 &
              0.24 & 0.19 & 0.18 &
              0.36 & 0.27 & 0.28 &
              0.0 & 0.0 & 0.0\\
 &Prophetnet&0.23 & 0.14 & 0.19 &
              0.38 & 0.40 & 0.36 &
              0.42 & 0.35 & 0.38 &
              0.0&0.0&0.0\\
 \midrule
 \parbox[t]{2mm}{\multirow{4}{*}{\rotatebox[origin=c]{90}{\small Pipeline + ft}}}& BART &0.44 & 0.31 & 0.37 &
              0.52 & 0.48 & 0.49 &
              0.45 & 0.36 &0.40 &
              0.72 & 0.62 & 0.72 \\
  &T5& 0.36 & 0.22 & 0.28 &
              0.51 & 0.45 & 0.47 &
              0.42 & 0.35 & 0.38 &
              0.66 & 0.48 & 0.66\\
 &Pegasus & 0.41 & 0.35 & 0.34 &
              0.52 & 0.45 & 0.49&
              0.39 & 0.31 & 0.34&
              0.79 & 0.63 & 0.79\\
 &Prophetnet& 0.43 & 0.32 & 0.37 &
              0.53 & 0.44 & 0.48 &
              0.38 & 0.32 & 0.34 &
              0.42 & 0.24 & 0.42 \\

 \bottomrule
\end{tabular}
\end{adjustbox}

\label{tabSummaries}
\end{table*}

% % BART+pipeline
% Suggestion Gold Mean Inverse Compression/reduction 0.2 variance: 0.13
% Suggestion system Mean Inverse Compression/reduction 0.25 variance: 0.16
% Information Gold Mean Inverse Compression/reduction 0.19 variance: 0.12
% Information system Mean Inverse Compression/reduction 0.18 variance: 0.13
% Experience Gold Mean Inverse Compression/reduction 0.13 variance: 0.13
% Experience system Mean Inverse Compression/reduction 0.1 variance: 0.08
% Question Gold Mean Inverse Compression/reduction 0.04 variance: 0.04
% Question system Mean Inverse Compression/reduction 0.01 variance: 0.02

% #BART+Ans
% Suggestion Gold Mean Inverse Compression/reduction 0.2 variance: 0.13
% Suggestion system Mean Inverse Compression/reduction 0.12 variance: 0.08
% Information Gold Mean Inverse Compression/reduction 0.19 variance: 0.12
% Information system Mean Inverse Compression/reduction 0.12 variance: 0.07
% Experience Gold Mean Inverse Compression/reduction 0.13 variance: 0.13
% Experience system Mean Inverse Compression/reduction 0.05 variance: 0.03

\subsection{Human Evaluation of Summaries}
Following \citep{fabbri2021summeval,zhan2022you}, we conduct a human evaluation of the system-generated summaries from BART for both Ans+ft and Pipeline+ft strategies to assess the quality of summaries holistically.\footnote{We provide an example of question-answers, gold standard and system-generated summaries from these two BART variants in Appendix \ref{appendix:generated_summaries}.} For each aspect category, we randomly sample 2\textendash 3 summaries from the test set where that category is present in both the gold and the system summaries using both model variants. Three of the authors, having a background in text-summarization research, annotate 21 summaries on a Likert scale from 1 to 5 (1 being lowest) along the following dimensions: \\\textbf{Coherence}\textemdash How well-organized are all the sentences collectively? \\ \textbf{Consistency}\textemdash is the summary logically implied (entailed) by the source? Summaries that contain hallucinated facts are penalized.\\ \textbf{Fluency}\textemdash how well written is each sentence.  Formatting, capitalization, or grammatical errors degrading the readability are penalized. \\\textbf{Relevance}\textemdash The summary should include only relevant and non-redundant information from the source answers. We also penalize sentences irrelevant to the summary corresponding to a particular aspect. For example, an \textit{Experience} may not be relevant for the \textit{Suggestion} summary\\
\textbf{Coverage}\textemdash we add this dimension to assess how well a summary covers a particular aspect from the answers (if present). 

\begin{figure}[ht]
    \centering
\includegraphics[width=0.49\textwidth, height=6.5cm]{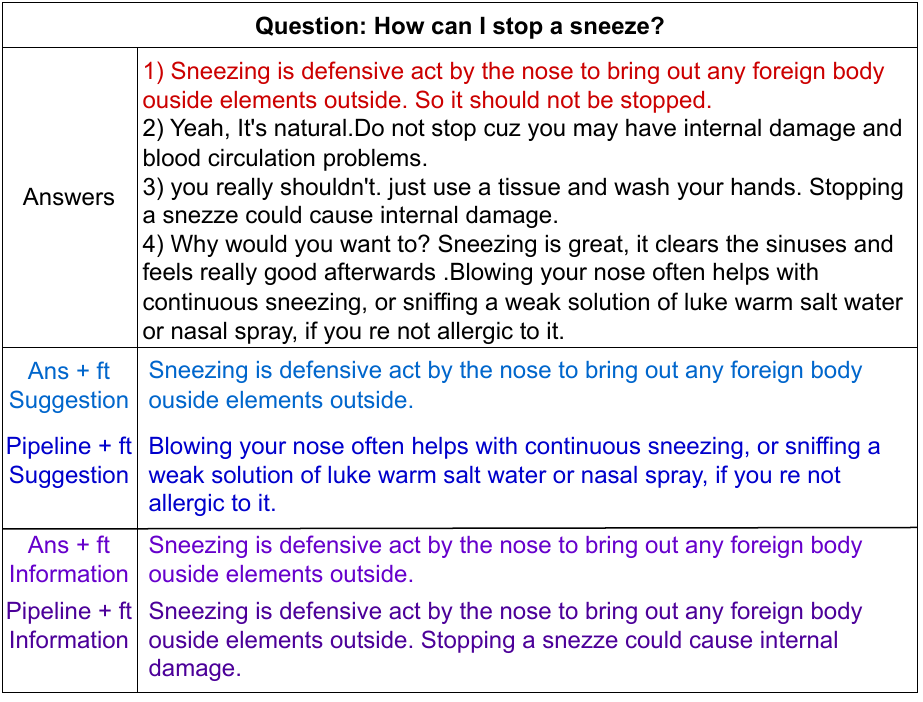}

    \setlength{\belowcaptionskip}{-15pt}\caption{Example summaries for \textit{Suggestion} and \textit{Information} generated by Ans + ft BART and Pipeline+ft BART models along with the best answer summary highlighted in \textcolor{red}{red}.
    }
    \label{fig:system_summary}
\end{figure}

We report the results in Table \ref{tab:human_Eval}. We note that the models have high coherence in general. We also observe that the summaries are quite extractive, which is why the models are scoring high in consistency. However, there is a slight drop for the Ans+ft version. Here, the model adds negation to one of the answers in the source document. Further inspection reveals that this hallucination might be grounded in real-world knowledge. We also note a lack of fluency, more so in the Pipeline+ft version. This, too, can be attributed to the extractive nature of generated summaries, where typos and grammatical disfluencies in the source text get transferred to the summaries as well. We can see this in Figure \ref{fig:system_summary} where the misspelled "ouside" and "snezze" are copied from the source answers to the summaries. Finally, Pipeline+ft summaries rank high in relevance and coverage in comparison to the Ans+ft strategy, as exemplified in Figure \ref{fig:system_summary}. This highlights the strengths of the pipeline in weeding out irrelevant sentences and preserving pertinent information at the same time. We quantify the disagreements between the annotators in terms of standard deviations across the scores and find high deviations in the average relevance scores. However, all annotators consistently score the pipeline summaries higher on relevance. We also see in Figure \ref{fig:system_summary} how the \textit{best answer} lacks coverage of an important aspect which is \textit{Suggestion}. This is addressed in the pipeline-generated summaries in this example.

\begin{table}[htbp]
\centering
\caption{Averaged human evaluation scores across three annotators. The standard deviations are reported in parenthesis} 
% \begin{adjustbox}{max width=.44\textwidth}
% \scriptsize
\resizebox{1\linewidth}{!}{%
\begin{tabular}{lccccc}
\toprule
\textbf{Model} & \textbf{Coherence} &\textbf{Consistency} &\textbf{Fluency} & \textbf{Relevance} & \textbf{Coverage}  \\ \midrule
\multirow{2}{*}{BART Ans+ft} & 4.81	& 4.44 &4.52&	3.11&	2.67\\
&(0.32)& (0.29)& (0.06)& (0.78)& (0.29)\\
\multirow{2}{*}{BART Pipeline+ft }& 4.76& 5.00	& 4.00 & 4.39 &4.48\\
&(0.05)&(0.00)&(0.09)&(0.34)&(0.14)\\
\bottomrule
\end{tabular}
}
% \end{adjustbox}
 
\label{tab:human_Eval}
\vspace{-1em}
\end{table}

% scale=0.53,keepaspectratio
% width=0.4\textwidth

\section{Conclusion} 
% Additionally, the question-answer threads in these platforms afford valuable insights into the prevailing public perceptions on specific health-related matters \citep{Odlum2015WhatCW}. how do our summaries help??
Using a multi-stage annotation framework, we introduce a novel problem of aspect-based health answer summarization from a community question-answering platform. This can be especially useful in low healthcare access settings for people seeking answers to common concerns or support from other users who might have experienced similar problems without being lost in a flood of answers on the public forums. We hypothesize that aspect-driven summaries can also improve performance on downstream question-answering tasks by separating information and suggestions from questions and subjective experiences. We provide an automated summarization pipeline built on our dataset using the latest state-of-the-art transformer-based models. The human evaluation demonstrates that our pipeline significantly improves over directly summarizing source answers. We also note that our linguistics heuristics-informed feature-based model offers a good baseline for aspect type classification. Future works can leverage these heuristics to construct a larger silver standard dataset to pre-train the aspect classifier before fine-tuning it on the gold annotations. In the future, our dataset can also be further augmented with factuality scores for each sentence with the help of an expert-sourced website. The summaries could then be rewritten to flag the sentences in \textit{Information} and \textit{Suggestion} categories with low expert confidence. % other perspectives such as Emotional Support As a future work, the data could be further augmented 

\section*{Limitations}
Summarization is a complex task for humans and, more so, for systems. We observed some limitations while working with our dataset and the proposed pipeline. Due to the pipeline approach, the errors from previous steps also hurt the performance of subsequent steps. For instance, failing to capture a relevant sentence reduces classification and summarization performance. Further, the human evaluation revealed consistency issues that could arise with a modern abstractive summarization system. Due to the informal nature of the source platform, we also notice several disfluencies in the summaries propagated from the source answers. 

\section*{Ethics Statement}
Since our work involves user-generated answers to health-related queries, we suggest caution and remind the readers that these summaries should not be treated as medical advice. While there is a chance they may propagate misinformation from the source answers, previous studies have found this chance to be a small one, usually negated by other users \citep{brady2016you}. Public health platforms are generally believed to empower individuals \citep{pitts2004illness} and contribute to complementing rather than substituting traditional health services \citep{kivits2006informed}. The purpose of our research is to facilitate access to information including but not limited to modern medicine, home remedies, or alternative treatments, as well as discussions of specific health-related products, information regarding particular health institutions, and financial or public health concerns across the globe or in a specific geographical context. Despite being generated from a public forum, the Yahoo!Answers data is entirely de-identified, alleviating the concern of releasing sensitive personal health information.
\balance
\section{Bibliographical References}\label{sec:reference}
\balance
\bibliographystyle{lrec-coling2024-natbib}
\bibliography{main}

\appendix
\onecolumn
\centering
\setcounter{page}{1}
{\textbf{{\Huge{Appendix}}}}
\section{Aspect Types Description}
\label{appendix:perspectives}
\begin{table*}[htbp]
    \centering
    \caption{Definitions and Examples of answer aspect types. Examples comprise a question and an answer sentence with the given aspect.}
    \resizebox{1\linewidth}{!}{
    \begin{tabular}{p{2cm}p{15cm}p{10cm}}
    \toprule
   \textbf{Aspect}& \textbf{Description}&\textbf{Example:Question, Answer Aspect} \\
   % \textbf & &Question, Answer Perspective\\
   \midrule
% \begin{enumerate}
\textit{Treatment}& A medical or therapeutic intervention aimed at addressing the health concern. &\vspace{-\abovedisplayskip}\smallskip\begin{minipage}{\linewidth}  Q. How can I relieve TENDONITIS in my shoulder??	\\A. Your doctor can give you a cortisone injection directly to the site of the pain to decrease the inflammation.\end{minipage}
\\\midrule

\textit{Suggestion}& A recommendation or advice such as home remedies, referral to a care provider, etc. Aimed at preventing a health issue from arising or exacerbating or to alleviate the symptoms. & \vspace{-\abovedisplayskip}\smallskip\begin{minipage}{\linewidth}  Q. um can anyone tell me y there's white stuff on my face?	\\A.it could be dry skin so switch to a sensitive moisturizer.\end{minipage}\\\midrule

  \textit{Experience}& People often share their personal experiences in support of the query. These may contain valuable subjective insights about what did or did not work for the people who have been in the same situation. Perhaps this is the most important aspect which attracts people to public forums for their health related concerns. & \vspace{-\abovedisplayskip}\smallskip\begin{minipage}{\linewidth}  Q. Newly diagnosed with sleep apnea - please advice!?	\\A. mine is treated with a machine which i wear while i sleep.\end{minipage} \\\midrule
  
 \textit{Information}& Any general knowledge related to various aspects of healthcare, such as epidemiology of a disease, whether a disease is progressive, how effective a certain medicine is, what is entailed in a particular medical procedure, general statistics about public health issues in a particular country etc. & \vspace{-\abovedisplayskip}\smallskip\begin{minipage}{\linewidth} Q. what is coumadin?	\\A. It is used for the prophylaxis of thrombosis and embolism in many disorders.\end{minipage}\\\midrule
 
\textit{Cause}& Information regarding the root cause of a problem or possible medical issues underlying certain symptoms. & \vspace{-\abovedisplayskip}\smallskip\begin{minipage}{\linewidth}  Q. what can cause liver enzimes to be elevated? \\A. Infection, such as viral hepatitis and mononucleosis.\end{minipage}\\\midrule

 \textit{Complications}& This category is to alert the answer seeker regarding possible expected or unexpected side effects of medications or procedures or adverse conditions that may arise as a result of pre-existing medical conditions. Sometimes these can be easily managed or can be severe and life threatening in some cases. However, it is important to have cognisance of these for proper decision making (whether to go for a treatment or not), and to monitor the possibilities for prompt recognition and timely management. &\vspace{-\abovedisplayskip}\smallskip\begin{minipage}{\linewidth} Q. Is asthma a progressive illness? Will it continue to get worse, damaging the lungs and become worse?	\\A. You can get permanent scarring in your lungs if it is not managed well. \end{minipage}
 \\ \midrule
\textit{Question} & In several cases the original question may miss out important details that are necessary to accurately answer it. In such cases, the answer providers may seek further clarification. &\vspace{-\abovedisplayskip}\smallskip\begin{minipage}{\linewidth}  Q. How can I find one of the best neurologists in the United States?	\\A. What is your condition?\end{minipage}\\\midrule
\textit{Emotional-Support} & Words of comfort to address the emotional and psychological needs of a patient or their loved ones. This can have significant positive impact on a person's coping mechanism during challenging times. A reassurance, and validation of person's feeling can bring a lot of comfort and may help reduce stress and anxiety during the time of distress. &\vspace{-\abovedisplayskip}\smallskip\begin{minipage}{\linewidth}  Q. anyone know of a good cookbook for kids with food allergies?	\\A. I'm so sorry about your son's allergies.	\end{minipage}
\\
% \end{enumerate}
\bottomrule
 \end{tabular}}

 \label{tab:perspectivedefinition}
\end{table*}

% \clearpage
\section{Subcategory Statistics}
\label{appendix:data}

\begin{figure*}[htbp]
\centering
\begin{minipage}{0.49\textwidth}
    \includegraphics[width=\linewidth]{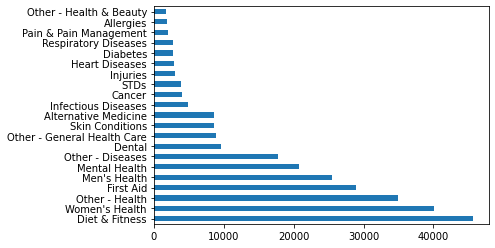} 
    \subcaption{Distribution of categories}
        \end{minipage}
    \begin{minipage}{0.49\textwidth}
    
    \includegraphics[width=\linewidth]{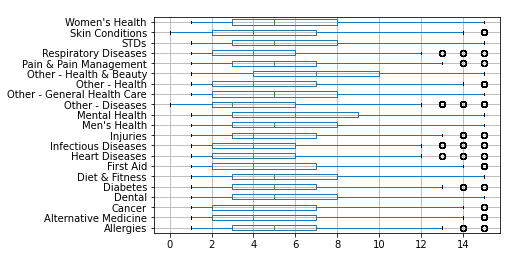}
\subcaption{Answers distribution across categories}
        \end{minipage}
\caption{Subcategory statistics. Panel (a) shows the distribution of threads across sub-categories. While Panel (b) shows the distribution of number of answers per post across various sub-categories} \label{fig:subcat-distribution}
\end{figure*}

\clearpage
\section{Preprocessing}
\label{appendix:preprocessing}
For creating sentence-level dataset, we apply a number of text cleaning and preprocessing steps.
\begin{enumerate}
    \item If a sentence doesn't end with .,! or ? and the next sentence starts on a new line with a lower case letter, we join the two. Otherwise, add a full stop after the first sentence. 
    \item We also add a full stop at the end of the paragraph if there's no punctuation marking end (!?.). 
    \item We remove punctuation symbols ($\sim$*()><"':-), URLs, smileys, excess newlines, and whitespaces, and replace repeated punctuations such as ..., ??, !!!, with single punctuation symbol. 
    \item We then tokenize all sentences using NLTK sentence tokenizer. 
    \item We find that the tokenizer misses out on some nuances, and we fix them by splitting the tokenized sentences again on a full stop. We make sure not to split on abbreviations such as etc., M.D., dr., i.e., P.S., L.A. or numerals such as 1., . etc. 
    \item After splitting, if a sentence contains less than two alphanumeric characters, we exclude it. 
    \item We again ensure a full stop at the end of each sentence that doesn't end in one of !, ? or . and remove extra whitespaces.
\end{enumerate}

\section{Average Compression Across Health Categories}
\label{appendix:compression_health_cat}

\begin{figure*}[ht]
    \centering
    \includegraphics[width=.9\textwidth]{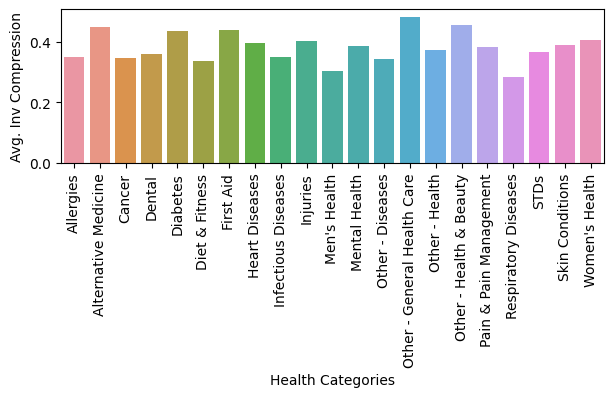}
    \caption{Average Reduction Across Health Categories. }
    \label{fig:compression_cat}
\end{figure*}

\clearpage
\section{Implementation Details}\label{appendix:implementation}
\subsection{Relevant Sentence Selection}
% SBERT encodes each sentence in a pair of sentences via a BERT-based encoder separately. Then a pooling is applied over it followed by a feed-forward layer to classify sentences as related or unrelated. The objective is to project related sentences close together in the embedding space and separate unrelated sentences. Results for `S-PubMedBert-MS-MARCO', `all-miniLM-L6-v2', `nli-roberta-base', `distilroberta-base-sentence-transformer' are available on request.
% \subsection{Perspective Type Identification}
% we also fine-tune distilroberta which has worse performance.
\subsection{Moods Classifier}
\label{appendix:moodsclf}
The model is trained on data combined of questions from SQuAD 2.0 dataset, imperatives, declaratives, Yes/No and Wh-questions from SPAADIA corpus. It is further enriched with manually written imperative statements. The final data has 1264 imperatives, 81104 indicatives, and 131219 interrogatives. This data is split into  training and validation set in 80:20 ratio. The input text is augmented with Parts-Of-Speech (POS) tags in the following way: word POS\_tag word POS\_tag. A Convolutional Neural Network classifier is trained on this data with max-pooling over time. The model is implemented with Keras API and uses 1-D convolution layer with 100 filters and kernel size 5 followed by global max pooling. The output is passed through a hidden layer with 500 hidden units and relu activation before being passed through a final dense layer with 3 units and softmax activation. A dropout of 0.2 is applied after embedding and hidden layers. The model is trained for two epochs with batch size 64. We add class-weights to the model to address the class imbalance and improve the performance across the rare `imperatives' class, bringing the macro-F1 score to 98.94\%
% \end{itemize}

\subsection{Question Features}
\label{Appendix:ques_features}
 \paragraph{do\_Patterns} "do i", "do you", "do they", "does she", "does he", "would you","is there", "are there", "is it so", "is this true", "are you", "is he", "is she", "is that true", "are we", "am i", "question is", "tell me more", "can i", "can we", "tell me", "can you explain", "to ask"
 \paragraph{helping\_Verbs} "is", "am","can", "are", "does", "would", "could", "can", "will"
 
 % \paragraph{Question Classifier}
 % The training data for this classifier is NPS Chat Corpus containing more than 10K threads from various sources tagged with several dialogue acts including Wh-questions and Yes/No questions. We modify the labels to binary indicators for question (True if Wh/Yes/No-Questions, false otherwise). We divide the data into training and validation splits in the ratio 80:20 and train a GBM classifier with 400 estimators using word-level Tf-IDF features (1-3 ngrams) and achieve a macro-average $F_1$ score of 87.25 (a Logistic Regression model with regularization gives us an $F_1$ score of 76.87) 
 
\subsection{Summary generation}
\label{appendix:summary generation}
We fine-tune a total of $32$ models\textendash$4$ summarization models and $2$ experimental scenarios (as described above) for the $4$ aspects. We use the pre-trained variants \textit{ainize/bart-base-cnn}, \textit{t5-small}, \textit{microsoft/prophetnet-large-uncased}, and \textit{google/pegasus-large} from Huggingface. perform manual hyperparameter tuning and save the model with the best performance on the validation set. We train the models for 5\textendash10 epochs, using batch sizes of 4\textendash16 and learning rate 1e-5\textendash3e-5 with a weight decay of 0.1\textendash0.2.

\clearpage
\section{Example of Aspect-based Health Answer Summaries}
\label{appendix:generated_summaries}  
Table \ref{table:qa_example} contains an example instance of a question and available answers from the data. Table \ref{table:generated_summaries} contains gold and system-generated summaries from two variants of the BART model.
\begin{table*}[!ht] 
\centering
\caption{An instance of question-answers from the data.}
% \begin{adjustbox} {max width=0.9\textwidth}
\small
\begin{tabular}{p{1.5cm} p{1.2cm} p{12.2cm}}
\toprule
\multirow{3}{*}{\textbf{Question}} & \textbf{Subject} & Spotty Monday Morning?\\
& \multirow{2}{*}{\textbf{Content}}& Why have I broke out in so many spots today?  Can this happen from lack of sleep?  Is that why its called beauty sleep?\\\hline
\multirow{21}{*}{\textbf{Answers}} & $1$.(Best Answer) & lack of sleep can cause lots of disorders and even blemishes. i think they originally called it "beauty sleep" as a pun from the fairy tale "sleeping beauty". but in real life, lots of sleep does promote healthier skin and well-being.\\
& $2$. & it could be lack of sleep, or it could be due to if you had a night out last night, especially if you were in a smoky pub and if you were drinking. if not it could be signs of the start of a cold. try washing your face with an anti bacterial face wash before you go to bed and when you get up in the morning, you should notice a difference after a few days\\
& $3$. &Have been depressed lately or on your period? Stress and PMS cause an increase in hormones which causes the human body to imitate what it was like to be 14 or 15. Another reason could be dairy, if you have recently had a lot of milk, cheese, ice cream, or chocolate, that could be why. Dairy speeds up our oil secretion in the same fashion as the stress. Try drinking at least 16 ounces of water a day it helps flush out things like extra oil in our bodies.College Anatomy and Physiology was a big helper in this one.  Hope this helped,Good luck.\\
& $4$. & lack of sleep doesn't help - but i would have thought that would be caused by a few nights bad sleep,  not just one.  my skin tends to be better during certain times of the month.  maybe it's due to hormones ?  try washing your skin with a cleanser b4 u go to bed,  and treat the really bad ones with either tee tree oil (tiny dab on the spot)  or put toothpaste on them overnight - just don't forget to wash it off in the morning !!!  toothpaste is excellent at helping to shrink spots.  it fights germs, and drys out excess oils.  just don't put it near your eyes though,  it'll make your eyes sting,  even if it's just close to your eyes....\\
& $5$. & No you dont get spots from lack of sleep.  You get them from what you eat/drink and from your skin type.  Make sure you exfoliate twice a week to remove all the dead skin cells and pop a couple of Vit E pills/day these are excellent for your complexion.
\\
& $6$. & Tell me about it!! its horrible isnt it!!
I definatly think lack of sleep and alcohol takes its toll on your skin!It definatly does mine!I think the best thing to do is if you do have too much to drink make sure you you keep your self hydrated the following day with plenty of water!! But yep i think its true what they say about needing your beauty sleep!!I certinaly wish id had mine this weekend then maybe i wouldnt look like this!!\\

\bottomrule
\end{tabular}
% \end{adjustbox}
\label{table:qa_example}
\end{table*}

\begin{table*}[!ht] 
\centering
\caption{Aspect-based gold summaries, and system-generated abstractive summaries.}
% \begin{adjustbox} {max width=0.9\textwidth}
% {|p{1.4cm}|p{1.6cm}|p{5.4cm}|p{8cm}|}
\small
\begin{tabular}{p{1.4cm} p{1.5cm} p{12cm}}
\toprule
\textbf{Summary Type}& \textbf{Aspect}&\textbf{Summary}\\
\hline
\multirow{21}{*}{ \parbox[t]{1.3cm}{\textbf{Gold Extractive}}} & \textit{Suggestion} &try washing your face with an anti bacterial face wash before you go to bed and when you get up in the morning, you should notice a difference after a few days
Try drinking at least 16 ounces of water a day it helps flush out things like extra oil in our bodies.try washing your skin with a cleanser b4 u go to bed,  and treat the really bad ones with either tee tree oil (tiny dab on the spot)  or put toothpaste on them overnight - just don't forget to wash it off in the morning !!!  toothpaste is excellent at helping to shrink spots.  it fights germs, and drys out excess oils. Make sure you exfoliate twice a week to remove all the dead skin cells and pop a couple of Vit E pills/day these are excellent for your complexion. I think the best thing to do is if you do have too much to drink make sure you you keep your self hydrated the following day with plenty of water!!\\
&\textit{Experience} & my skin tends to be better during certain times of the month. I definatly think lack of sleep and alcohol takes its toll on your skin!It definatly does mine! But yep i think its true what they say about needing your beauty sleep!!I certinaly wish id had mine this weekend then maybe i wouldnt look like this!!\\
&\textit{Information} & lack of sleep can cause lots of disorders and even blemishes. i think they originally called it "beauty sleep" as a pun from the fairy tale "sleeping beauty". but in real life, lots of sleep does promote healthier skin and well-being.
it could be lack of sleep, or it could be due to if you had a night out last night, especially if you were in a smoky pub and if you were drinking. if not it could be signs of the start of a cold. Stress and PMS cause an increase in hormones which causes the human body to imitate what it was like to be 14 or 15. Another reason could be dairy, if you have recently had a lot of milk, cheese, ice cream, or chocolate, that could be why. Dairy speeds up our oil secretion in the same fashion as the stress. lack of sleep doesn't help - but i would have thought that would be caused by a few nights bad sleep,  not just one. maybe it's due to hormones ? No you dont get spots from lack of sleep.  You get them from what you eat/drink and from your skin type.\\
&\textit{Question} & Have been depressed lately or on your period? \\\hline
% & & \\
\multirow{11}{*}{ \parbox[t]{1.3cm}{\textbf{Gold Abstractive}}} &\textit{Suggestion} & Wash your face with an anti bacterial face wash before sleeping and in morning and drink lots of water to flush out extra oil from body. Really bad ones can be treated with either tee tree oil (tiny dab on the spot)  or toothpaste overnight to fights germs and dry out excess oils. Exfoliate twice a week to remove all the dead skin cells and take Vit E pills/day.\\
&\textit{Experience} & my skin tends to be better during certain times of the month. Lack of sleep and alcohol takes its toll on my skin. I wish I had a good sleep this weekend so I looked better.\\
&\textit{Information} & lack of sleep can cause lots of disorders and even blemishes. It could also be if you spent night out in a smoky pub and alcohol is another cause. It could also be a signs of the start of a cold. Stress and PMS cause an increase in hormones which could be a possible cause as well. Another reason could be dairy which speeds up our oil secretion. You get them from what you eat/drink and from your skin type.\\
&\textit{Question} & Have been depressed lately or on your period?\\
\midrule
\multirow{5}{*}{ \parbox[t]{1.3cm}{\textbf{BART Ans + ft:}}}& \textit{Suggestion} & If you want to look at someone who has beaten stage 4, then look at Lance Armstrong. He is the poster of miracles.\\

&\textit{Information} & It must be very difficult to see your friend go through all of this.\\

&\textit{Experience} & I read a book by a lady who did via the Gerson therapy.\\

&\textit{Question} & (No summary generated)\\
\midrule 
\multirow{14}{*}{\parbox[t]{1.3cm}{\textbf{BART Pipeline + ft:}}}
& \textit{Suggestion} & Try washing your face with an anti bacterial face wash before you go to bed and when you get up in the morning, you should notice a difference after a few days. Try drinking at least 16 ounces of water a day it helps flush out things like extra oil in our bodies.\\
&\textit{Experience} & My skin tends to be better during certain times of the month.. I definatly think lack of sleep and alcohol takes its toll on your skin!\\
&\textit{Information} & lack of sleep can cause lots of disorders and even blemishes, but in real life, lots of sleep does promote healthier skin and well being.. it could be due to if you had a night out last night, especially if you were in a smoky pub and if you are drinking.. if not it could also be signs of the start of a cold.. Stress and PMS cause an increase in hormones which causes the human body to imitate what it was like to be 14 or 15.. Another reason could be dairy, if you have recently had a lot of milk, cheese, ice cream, or chocolate, that could be why.. Dairy speeds up our oil secretion in the same fashion as the stress.. lack of sleep doesn't help but i would have thought that would be caused by a few nights bad sleep, not just one.. No you dont get spots from lack of sleeping.. You get them from what you eat/drink and from your skin type.\\
&\textit{Question} & Have been depressed lately or on your period? Maybe it s due to hormones?\\
\bottomrule
\end{tabular}
% \end{adjustbox}

\label{table:generated_summaries}

\end{table*}
\end{document}